\documentclass[10pt,conference,a4paper]{IEEEtran}

\usepackage{color}
\usepackage{graphicx}
\usepackage{subfig}

\begin{document}
\title{With Whom Do I Interact? \\ Detecting  Social Interactions in Egocentric Photo-streams}

\author{\IEEEauthorblockN{Maedeh Aghaei}
\IEEEauthorblockA{
University of Barcelona\\
Barcelona, Spain\\
Email:maghaeigavari@ub.edu}
\and
\IEEEauthorblockN{Mariella Dimiccoli}
\IEEEauthorblockA{
Computer Vision Center and\\ University of Barcelona\\
Barcelona, Spain}
\and
\IEEEauthorblockN{Petia Radeva}
\IEEEauthorblockA{
Computer Vision Center and\\ University of Barcelona\\
Barcelona, Spain}
}

% make the title area
\maketitle

% As a general rule, do not put math, special symbols or citations in the abstract
\begin{abstract}
Given a user wearing a low frame rate wearable camera during a day, this work aims to automatically detect the moments when the user gets engaged into a social interaction solely by reviewing the automatically captured photos by the worn camera. The proposed method, inspired by the sociological concept of F-formation, exploits distance and orientation of the appearing individuals -with respect to the user- in the scene from a bird-view perspective. As a result, the interaction pattern over the sequence can be understood as a two-dimensional time series that corresponds to the temporal evolution of the distance and orientation features over time. A Long-Short Term Memory-based Recurrent Neural Network is then trained to classify each time series. Experimental evaluation over a dataset of 30.000 images has shown promising results on the proposed method for social interaction detection in egocentric photo-streams.
\end{abstract}

% no keywords

\IEEEpeerreviewmaketitle

\section{Introduction}

Microsociology, or social interaction, as defined by Erving Goffman \cite{goffman} is a process by which people act and react to those around them. Social interaction is one of the most important factor in predicting the physical and mental health and well-being of everyone ranging from childhood through older adults \cite{healthsocial}. As a consequence, automatic recognition of social interactions from images and videos has increasingly drawn scientific interest \cite{surveysocial2}. Nonetheless, in order to obtain a rigorous study of the social interaction patterns from a subject, long term life observations from his point of view are required. In this regard, high frame rate wearable cameras (e.g. GoPro or Looxcie) are not useful, since they are not able to capture the whole day of the person. Instead, low frame rate (2fpm) wearable cameras (e.g. Narrative) are to be used to record the life of a person -including their social activities- from a first-person point of view for long periods of time. 

Several works focused on the automatic extraction of information from videos to describe their semantic content, have been developed before \cite{action,semanticextraction1,semanticextraction2}. As a specific type, different works have been dedicated to automatic recognition and understanding of social interactions in videos employing combination of information captured by different sensors such as camera, microphone, accelerometer, bluetooth, infrared and etc. \cite{salsa,surveysocial}. However, defining an interaction, relying solely on visual cues is a valuable task from computer vision perspective that confines the analysis to the visual information, eliminating the need for acquiring additional information and major privacy concerns. Distinct past efforts have been done in computer vision  to solve the social interaction detection problem in both normal and egocentric videos employing different visual features \cite{socialfathi,socialalletto,socialryoo,cristani2011social,surveycristani}. 
\begin{figure}% figure*
\includegraphics[width=\linewidth]{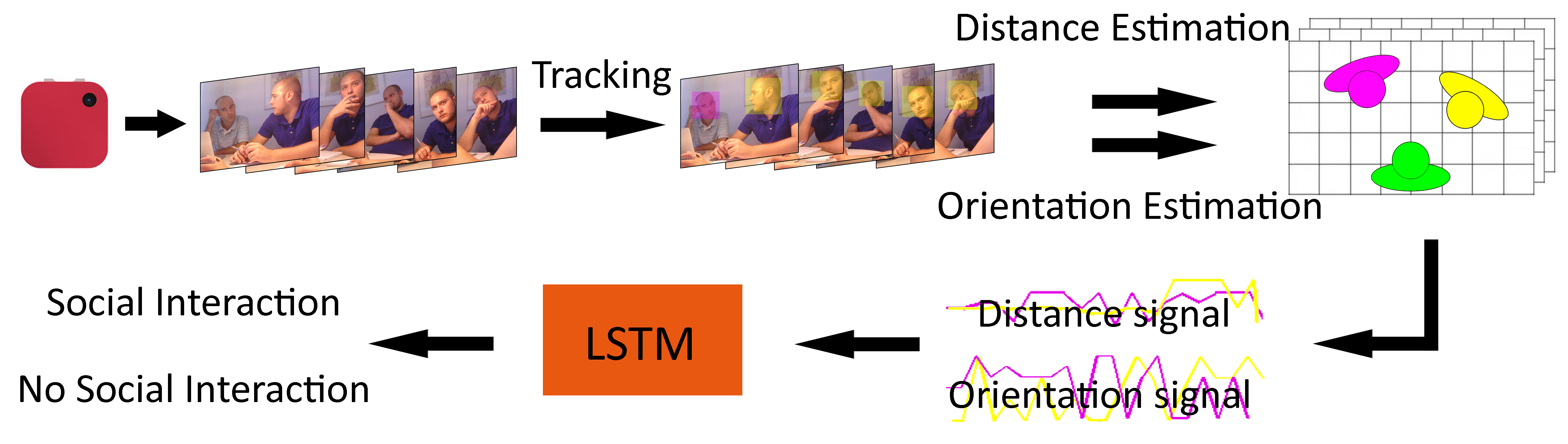}
\caption{Work flow of the proposed method.}
\label{methodology}
\end{figure}

When humans get involved in social interactions, they tend to stand in determined close positions to other interacting people to avoid occlusions and organize orientations in order to naturally place the focus on the subjects of interest. This phenomenon was first studied and described by Kendon in the theory of F-formation \cite{kendon}. F-formation is defined as a pattern that people instinctively maintain when interacting and can be measured based on the mutual distances and orientations of the subjects in the scene. F-formation comprises of 3 spaces: the people involved in an interaction stand in the p-space, where they all look inwards to a common empty space surrounded by the p-space that forms the o-space. External people who do not belong to this interaction are not accepted in the p-space and they belong to any space outside of the p-space known as the r-space.

\begin{figure}
\includegraphics[width=\linewidth]{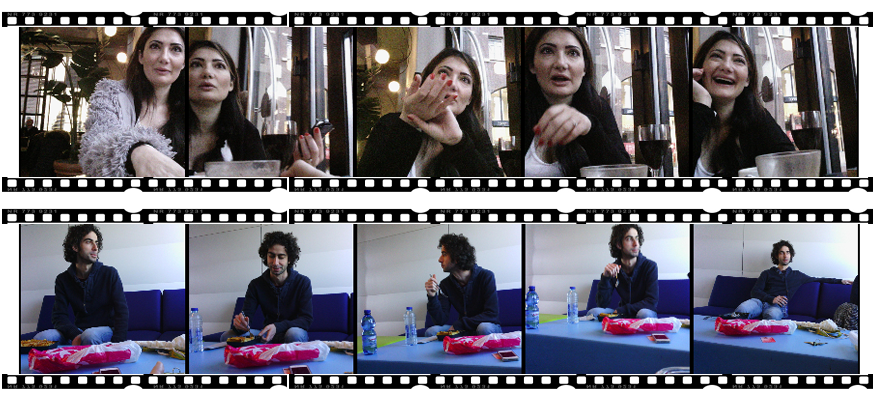}
\caption{Example of the camera wearer being socially interacting (first row) and not interacting (second row).}
\label{intvsnoint}
\end{figure}

%related work
%Social interaction detection in images and videos
Most of the previous works in social interaction computing were focused on finding potential groups of interacting people. Cristani et al. \cite{cristani2011social} introduced a framework, namely HVFF, to find F-formations in conventional videos that takes as input the positions of the people in a scene and their head orientations. To recognize the F-formation in every frame of a crowded scenario, the authors applied a voting strategy based on the Hough transform. Meanwhile, in the case of egocentric videos, their unique properties allow completely new approaches to social analysis. These videos at the same time pose new challenges to conventional methods such as strong ego-motion, background clutter, and severe changes in lighting conditions. Previous methods for social interaction detection in egocentric videos essentially were designed to address the problem on head mounted cameras videos with high frame rate. Fathi et al. in a pioneering work \cite{socialfathi}, presented a model which aims to classify social interactions into discussion, monologue and dialogues. In the presented framework, location and orientation of faces are estimated and used to compute the line of sight for each face and to estimate the 3D location they are looking at. A multi-label Hidden Conditional Random Field model is then used to assign a category to each social situation in the video. Recently, Alletto et al. \cite{socialalletto}, following the notion of F-formation proposed a model to organize people in the scene into socially related groups. In order to describe how two people stand in relation to one another, people are tracked in the video and their head pose and 3D location are predicted using an estimation framework exclusively developed for egocentric videos. Later, a correlation clustering algorithm is applied to merge pairs of people into socially related groups and finally, due to different definition that distances and orientations can have, a Structural SVM based approach is used to learn the weight vector of the correlation clustering depending on the occurring social situation.

Despite the aforementioned methods, our proposed approach tackles the problem of social interaction detection in photo-streams captured by a low frame rate wearable camera (2fpm). Our focus on real-world social events, such as coffee breaks, casual work meetings or a sudden encounter in a park, makes the task especially challenging due to the complex visual appearance of natural scenes and the presence of large numbers of individuals in addition to the social group of interest. Moreover, the sparsity of the observations and the camera being worn as part of clothing, lead to unpredictable changes in the field of view even from frame to frame. In this regard, we previously proposed an approach \cite{aghaei2015towards} to establish which individuals are more involved in an interaction with the camera wearer by analyzing pairwise interaction vector of the camera wearer with other individuals in every frame of a sequence employing the F-formation theory and adapting the HVFF strategy \cite{cristani2011social} to egocentric photo-stream scenario. The presence of a social interaction is decided in every single frame separately and eventually, if the number of found interactions with regard to the full length of the sequence is higher than a predefined threshold, then that specific person is considered as socially interactive. This work has two major drawbacks. First, the classification precision is highly dependent on the selected threshold for the task. Second, the inter dependency between frames and evolution of features over the time is not considered.

Despite most existing methods which make little use of the evolution of the features over time, in this work we employed Long Short-Term Memory Recurrent Neural Network (LSTM) which is adapted for learning over sequential data. The proposed method aims to describe how the evolution of the social clues characterizing the F-formation theory which inferred from human behavior, can be employed to decide if the appearing people in a sequence are interacting with the camera wearer or not. To the best of our knowledge, this work is first to detect social interactions with the camera wearer at sequence level instead of frame-to-frame level information in the domain of egocentric photo-streams.

%The rest of the paper is organized as follows, in section 2, we detail the proposed method, in section 3, we discuss experimental results and, in section 4, we give some concluding remarks.

\section{Methodology}

During the course of a day, people may engage in social events with specific persons, such as having lunch with family members. Due to their emotional impact, social events might be considered as special moments to be retrieved for the wearable camera user.
However, automatic social event recognition from images is not a trivial task. Although observing a single frames from the F-formation perspective leads to some coarse frame-based information about the status of the social interaction in that frame, but it also bears some uncertainty that makes the decision rather unreliable. To this end, sequence level analysis of the features to prove the involvement of the people in the social interaction with the camera wearer is required (see Fig. \ref{intvsnoint}).
Our proposal towards social interaction detection and classification in egocentric photo-streams comprises of two main modules. The first module aims to find the required descriptive features employing the F-formation for this task and prepares them in the sequence level. The second module analyzes the resulting features from the first module to classify the sequences. A visual overview of the proposed method is given in Fig. \ref{methodology}.

\subsection{Feature Extraction} \label{Feature extraction}
Following the F-formation notion to extract descriptive features, the method first localizes the appearance of each person in the scene along the photo-stream. Afterwards, pose and 3D location for each person along the sequence are estimated to build the set of features for the analysis. 
The features that are used for social interaction description and the methodology used to extract them are detailed in this section.

\subsubsection{People Localization in Photo-streams}

\begin{figure}
\includegraphics[width=\linewidth]{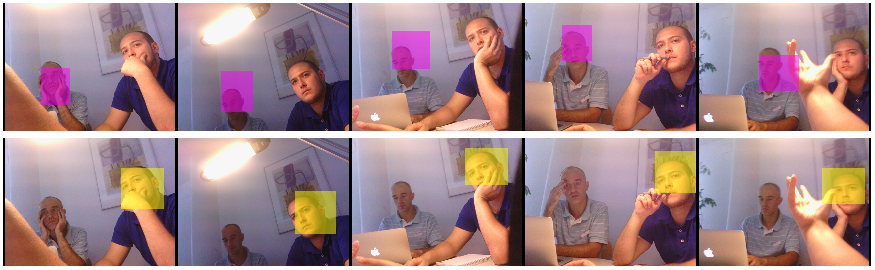}
\caption{Tracking result of applying eBoT \cite{ebot} over a sequence of two persons.}
\label{ebot}
\end{figure}

Tracking is undoubtedly a prior task for human behavior analysis. To detect and localize people around the camera wearer along a sequence, we use the multiple face assignment algorithm previously developed in egocentric photo-streams \cite{ebot}. Tracking aims to calculate the trajectory of every person in the scene and keep it across occlusions. We rely on faces instead of bodies as we believe faces and their attention patterns play more important and accurate role in determining a social interaction.

In the proposed tracking framework, the photo-stream is first segmented temporally into a set of sequences as a portion of the photo-stream in which the presence of the people (interacting or not) is established previously. In each sequence, for each visible face -known as seed- a tracklet is generated which is comprised of a set of correspondences along the whole sequence. Later, similar tracklets are pooled into the so called extended bag-of-tracklets (eBoT). All of the tracklets in an eBoT are aimed to track one specific person along the sequence, having the seed in different frames. Unreliable eBoTs are excluded from the original set of eBoTs by using a measure of confidence. After filtering reliable eBoTs, a prototype tracklet is extracted from them (see Fig. \ref{ebot}). Additionally, eBoT is able to estimate occluded frames from the final prototype and remove them. This eBoT property helps the social interaction analysis because the removed frames do not convey useful information and may bias the final classification results.

\subsubsection{Face Orientation Estimation}
Line of sight of a person can be roughly approximated by estimating the head pose of the person. To this end, for each detected face, we expand the bounding box around the face by a small factor and apply the state-of-the-art face pose estimation algorithm introduced by Zhu et al. \cite{facedetector} on the expanded region. The detector is based on mixture of trees with a shared pool of parts, where every facial landmark is defined as a part and a global mixture is used to model topological changes due to the viewpoint. This method is able to predict the head orientation among discretized viewpoint between -90$^{\circ}$ (looking to the left) to 90$^{\circ}$ (looking to the right) every 15$^{\circ}$ along pitch and yaw directions. The algorithm is being applied over a confined and relatively small area of the image. Hence, decreasing the detection threshold parameter increases the probability of truly estimating the head pose, while the probability of finding false positives remains relatively unchanged. This procedure is repeated for all the face regions. However, due to the {\em in the wild nature} of egocentric images, pose estimator fails to estimate the face orientation for some regions. Thus, for these cases, a pose relaying on the median poses of its neighboring frames is assigned.

\begin{figure}
\includegraphics[width=\linewidth]{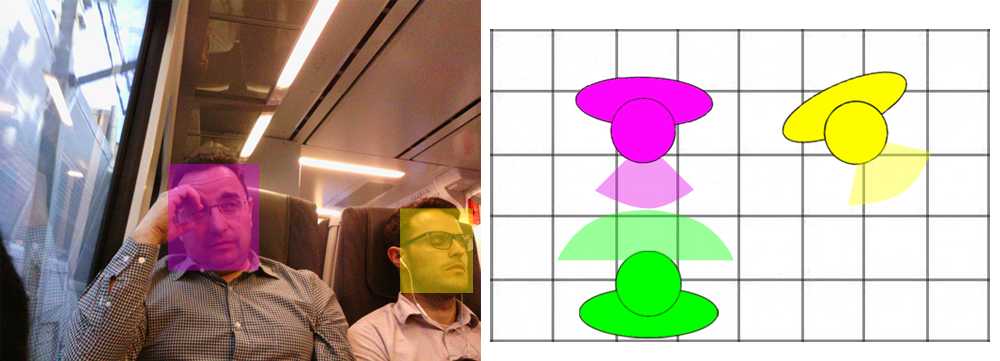}
\caption{On the right side, the bird-view model of the left side scene can be observed. The camera wearer is shown in green with an expanded line of sight region. Only the person depicted by pink color (left) is interacting with the camera wearer.}
\label{sightline}
\end{figure}

The line of sight of the camera wearer is not used explicitly in the analysis. However, it is important to know his line of sight to associate the line of sight of others to him. Note that the camera is typically worn on the chest of the camera wearer. Therefore, to predict his line of sight, we assume he can possibly look at anywhere from his left side to his right side, which results in 180$^{\circ}$ freedom (-90$^{\circ}$ to 90$^{\circ}$), while being more probable to look at the other people in the scene. On the other side, we assumed people to be interacting with the camera wearer should have a head pose between -30$^{\circ}$ to 30$^{\circ}$ (see Fig. \ref{sightline}).

\subsubsection{3D People Localization}
The F-formation model relies on a bird-view model of the scene, where each person is represented with two coordinates $(x,z)$, where $x$ denotes position of the person in the 2D image and $z$ denotes its distance to the camera. The $x$ position of the person is inferred from the $x$ position of the center of the bounding box which surrenders face of the person. 

To estimate the distance of each individual from the camera, a regression model that learns the depth relationships on a two-dimensional surface is trained. A second degree polynomial regression model learns the relation between the distance from the camera and the vertical height of a face. In this work, a distance of 150 cm (as the diameter of the o-space) is being assumed as the margin of a natural distance for a social interaction.

\subsection {Social Interaction Classification by LSTM}

The features described in previous subsection encode a local snapshot in time. However, the analysis of temporal change in these features is crucial for detection and understanding of social interactions. The next module of our proposed model employs these features for classification of the sequences into socially interacting or not. We model our problem using particular Recurrent Neural Network (RNN) classifier, namely LSTM, in order to take advantages of its ability to use the temporal evolution of the descriptors for classification. By introducing distance and orientation features of each person along a sequence, LSTM is able to decide which class the sequence pertains to. A brief description of LSTM and the architecture used in this work is provided in the following subsection.

\subsubsection{Introduction to LSTM}
As a special class of neural networks, RNN is able to capture the temporal dynamic of the sequences thanks to its recurrent connections between units being a directed cycle in the nodes. Nonetheless, the short term memory of RNN becomes insufficient when it needs to learn tasks that incorporate long time lags of unknown size between important events when  processing long sequences.
LSTM is a special type of the RNN capable of learning long term dependencies owing to its more powerful update equations. LSTM is able to do so by utilizing memory cells that use logistic and linear units with multiplicative interactions with input and output gates. In this way, it overcomes the \textit{exponential error decay} problem of RNN. LSTM has been tested in many applications ranging from language modeling to image captioning, and achieved remarkable results compared to existent methods for many tasks \cite{lstmdescribing,lstmkarpathy2015deep}.

\subsubsection{Classification by LSTM}
For egocentric sequence binary classification purpose, in this paper we propose to train a LSTM network by introducing to it the feature vectors extracted from each sequence as presented in section \ref{Feature extraction} at each time step. Both the distance and orientation information for each sequence are introduced to the network as input. The system must learn to classify sequences of different lengths to interacting or not by analyzing the two feature vectors associated to each sequence. Hence, the system needs to learn to protect memory cell contents against even minor internal state drift. 
%The sigmoid function ensures that the network outputs are between 0 and 1, at every time step.

The hidden layer contains several memory cells fully inter-connected and fully connected to the rest of the network. The input and output gates use inputs from other memory cells to decide whether to access certain information in its memory cell. Being the $i$th memory cell $c_i$, at time $t$, $c_i$'s output $y^{c_i} (t)$ is computed as:

$y^{c_i} (t) = y^{out_i}(t) h(s_{c_i} (t)),$

\noindent where the internal state $s_{c_i} (t)$ is 

$s_{c_i} (0) = 0$

$s_{c_i} (t) = s_{c_i} (t-1) +  y^{in_i}(t)g( net_{c_i} (t))$ for $t>0$

\noindent and being ${in_i}$ and ${out_i}$, cell's input gate and output gate, respectively. $g$ is a differentiable function that squashes $net_{c_i}$, and $h$ is a differentiable function that scales memory cell outputs computed from the internal state $s_{c_i}$. We refer readers to \cite{hochreiter1997long} for more detailed description of the LSTM architecture. Each unit of the input layer receives the distance and the orientation of a person  w.r.t the camera wearer in a given frame. For the output layer, a sigmoid function is used as activation function, which is standard for binary classification tasks.

Training of the LSTM network is essential for accurate classification of sequential inputs. In a simple feedforward network, backpropagation is a common method of training used in conjunction with an optimization method such as gradient descent. The method moves backward from the final error to assign responsibilities to the weights for a portion of the error by calculating their partial derivatives or the relationship between their rates of change. Those derivatives are then used by the learning method, to adjust the weights in a direction that decreases the error. 
%The method calculates the gradient of a loss function with respect to all the weights in the network. The gradient is fed to the optimization method which in turn uses it to update the weights, in an attempt to minimize the loss function.
For training a LSTM using Back Propagation Through Time (BPTT), a time element is added which extends the series of functions for which it calculates derivatives with the chain rule.

\section{Experiments}

There are two types of gradient-based techniques for LSTM training: truncated-BPTT \cite{hochreiter1997long} and full-BPTT \cite{graves2005framewise}. To validate our approach, we trained a LSTM with revised optimization technique introduced by Lyu and Zhu \cite{lstm-revisit}. The revised method incorporates full-BPTT under a matrix-based batch learning method where the overall performance for the LSTM is improved using revised activation functions for gates leading to sharper nodes.

\subsection{Dataset}

Due to the lack of public datasets with ground-truth information on wearable camera images, all the experiments presented in this paper were carried out on a novel egocentric dataset, where the presence or absence of interaction with each  person is manually labeled. The dataset was acquired by 8 users; each user wore the Narrative camera (2fpm) for a number of non-consecutive days over a total of $\sim$100 days period, collecting $\sim$30.000 images, where $\sim$3000 images of them contain a total number of 100 different trackable persons along sequences of average length of 25 frames. Sequences have different lengths, varying from 10 to 40 frames (5 minutes to 20 minutes of interaction). The dataset has been acquired in real world situations including inside and outside scenes and in different daylight conditions. 75\% of the dataset contains socially interacting persons and the rest 25\% of the dataset consists of not interacting persons.

To train a network for classification a relatively large amount of both socially interacting and non-interacting sequences are required. Note that a sequence for the network is defined by the two-dimensional times series (distance and orientation of each person along the sequence) extracted from each sequence. This amount of labeled data is not currently available and acquiring it is difficult and costly. However, augmenting the dataset is highly recommendable for classifiers training, when large representative training sets are not available \cite{synthetic}. Therefore, in this work we considered an extended training set resulted by augmentation of the available real dataset.

%Therefore in this work we considered both, a complete synthetic training dataset and a training dataset created by  randomly selecting out 70\% of our whole real dataset and artificially enlarged them in the feature space. In order to provide a balanced training set, 70\% of the dataset is chosen equally to form the training set from positive and negative sequences. 

The training dataset is created by randomly selecting out 70\% of our whole real dataset and artificially enlarging them in the feature space. In order to provide a balanced training set, 70\% of the real dataset is chosen equally from positive and negative sequences to form the training set. For each sequence in the real dataset, a set of repetitive sequences of randomly different length in the interval [10,40] is provided which randomly contains features of the original sequence. Later, to variate the sequences, a set of positive features and negative features (as to add some bias to the sequence) is introduced to the sequence. Positive features for socially interacting sequences are defined as ones with a random distance value in centimeters in the interval $[10,150]$ and a head pose orientation value in degrees in the interval  [-30$^{\circ}$,30$^{\circ}$]. Negative features for socially interactive sequences are just random values out of these boundaries, but in the valid boundary of these features. For the socially non-interactive sequences, these boundaries are just the opposite. Rest of the real dataset (30\%) is equally divided to form the validation and test set. We augmented the validation set in the same manner as for training set. Fig. \ref{real} shows the error and the correctly classified objects on the validation dataset.

\subsection{Network Architecture}

We use a 3-layer networks of one input layer with 2 input units, one hidden layer with memory cell blocks of size 2, and one output layer with 1 output unit. 
The input layer has forward connections to all units in the hidden layer. The fully self-connected hidden layer contains memory cells and corresponding gate units. The output layer receives connections only from memory cells. All activation functions are logistic with output range $[0,1]$, except for $h$, whose range is $[-1,1]$, and $g$, whose range is $[-2,2]$. All weights are randomly initialized in the range [$-0.1,0.1]$. We use peephole connections, but omit forget gates since this task is not a continual prediction task.

\subsection{Parameter Estimation}

The use of machine learning algorithms is tightly bounded with tuning of model hyperparameters. Due to the complexity of this task, there is  great appeal for automatic approaches that can optimize the performance of a given learning algorithm to the problem. Several sequential Bayesian Optimization methods have been proposed for hyperparameter search \cite{hutter2014efficient,eggensperger2015efficient}. The common theme is to perform a set of iterative hyperparameter optimization. These methods in each iteration fit a hyperparameter response surface that maps each hyperparameter setting to an approximated accuracy using a probabilistic regression function. The learned regression model is then used as a substitute of the response surface to quickly explore the search space and identify next hyperparameter candidates to evaluate.

In this work, we chose to optimize the hyperparameters of the LSTM separately using random search because this approach is easy to implement and parallelize, while covers the search space quite uniformly. We made a  total number of 3000 trials of randomly sampling over the following hyperparameters: log-uniform  samples in the interval  $[2,200]$ for the number of memory cells blocks,
log-uniform samples in the interval $[1e-5, 1]$ for the learning rate,
log-uniform samples in the interval  $[0.01, 1]$ for the momentum and, log-uniform samples in the interval  $[200, 1000]$ for the batch size.
We tried two different optimization techniques in the experiments:  one is Limited memory BFGS (L-BFGS) and the other one mini-batch Stochastic Gradient Descent (SGD), with different mini-batch sizes for LSTM \cite{ngiam2011optimization}.

\subsection{Experimental Results and Discussion}

In this work, we focus on the ability of LSTM for the classification problem.
In all the experiments, we trained the model on the real-based augmented training set and  used cross-validation to trigger early stopping. We used 10,000 training samples and  1,500 validation samples where the error on the validation set is used as a proxy for the generalization error in determining when overfitting has begun to stop training and use the weights the network had in that previous step as the result of the training run to test the model.

The unique characteristic of our scenario (the camera having low frame rate (2fpm), as well as being worn on the chest), makes the resulting method incompatible with most of the previously proposed models for social interaction classification to perform the comparison. However, We have compared the obtained results with our previously introduced work in the area which employs the HVFF to classify social interactions in the egocentric photo-stream scenario \cite{aghaei2015towards}. The obtained results using both methods can be observed in Table \ref{table1} which demonstrate the improvements on the results using the sequence-level analysis of the social interactions using LSTM.

\begin{table}
\centering
\begin{tabular}{c|cc|c|}
\cline{2-4}
&   \multicolumn{2}{c|}{\textbf{Our method}} & \textbf{HVFF}\cite{aghaei2015towards} \\ \cline{2-3}
 
&  \multicolumn{1}{c|}{\textbf{LBFGS}} & \multicolumn{1}{c|}{\textbf{SGD}} &  \\ \hline

\multicolumn{1}{|c|}{\textbf{Precision}} & \multicolumn{1}{c|}{82\%} & \multicolumn{1}{c|}{73\%} & \multicolumn{1}{c|}{80\%} \\ \hline
\multicolumn{1}{|c|}{\textbf{Recall}} & \multicolumn{1}{c|}{74\%} & \multicolumn{1}{c|}{85\%} & \multicolumn{1}{c|}{72\%} \\ \hline
\multicolumn{1}{|c|}{\textbf{F-measure}} & \multicolumn{1}{c|}{77\%} & \multicolumn{1}{c|}{78\%} & \multicolumn{1}{c|}{75\%} \\ \hline
\end{tabular}
\caption{The quantitative results of the proposed method as well as the comparative result with HVFF\cite{aghaei2015towards}.}
\label{table1}
\end{table}

% \begin{table}[]
% \centering
% \caption{My caption}
% \label{table1}
% \begin{tabular}{l|l|l|l|}
% \cline{2-4}
%                        & augmented & real & HVFF \\ \hline
% \multicolumn{1}{|l|}{Precision}  & 10 & 20 & 30 \\ \hline
% \multicolumn{1}{|l|}{Recall} & 10 & 20 & 30 \\ \hline
% \end{tabular}
% \end{table}
 
\begin{figure}
\centering
    \subfloat[]{{\includegraphics[width=4cm]{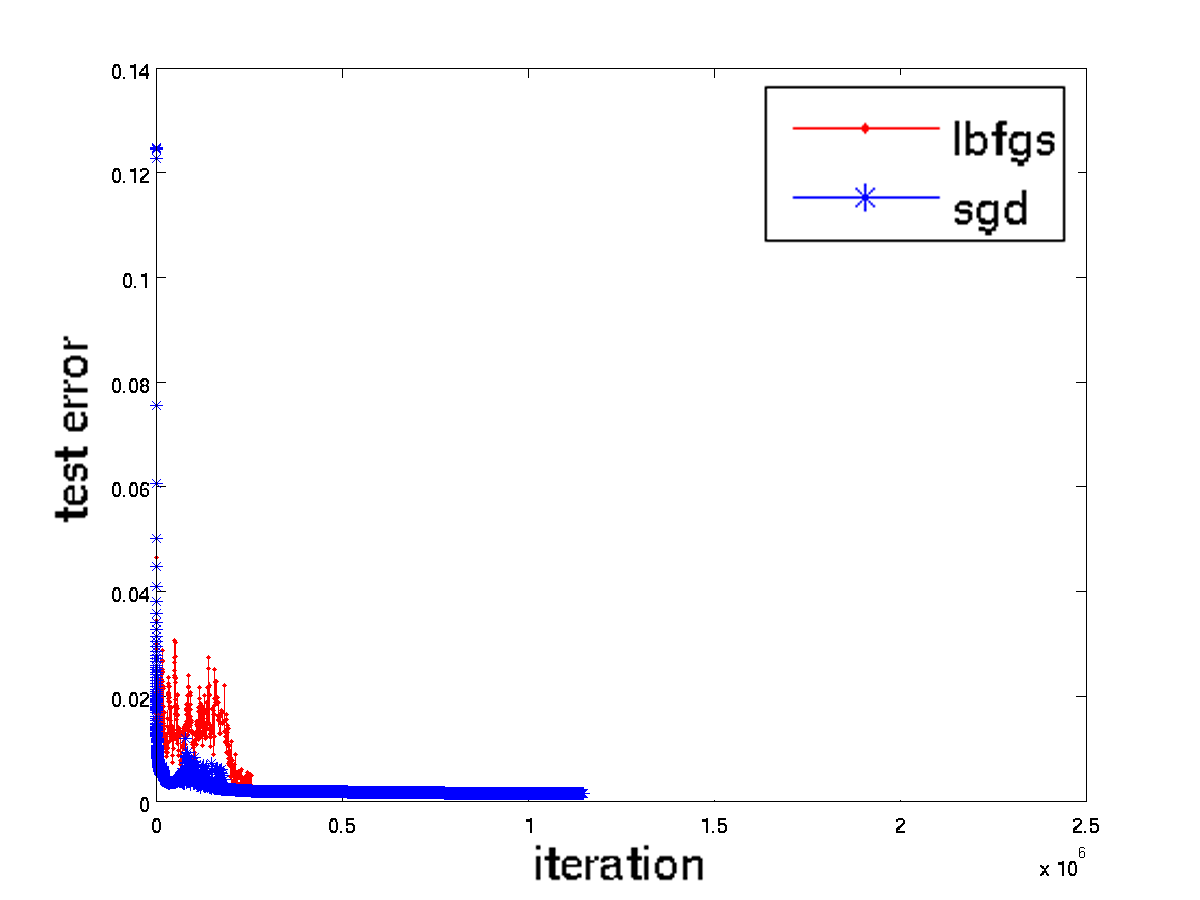} }}%
%     \qquad
    \subfloat[]{{\includegraphics[width=4cm]{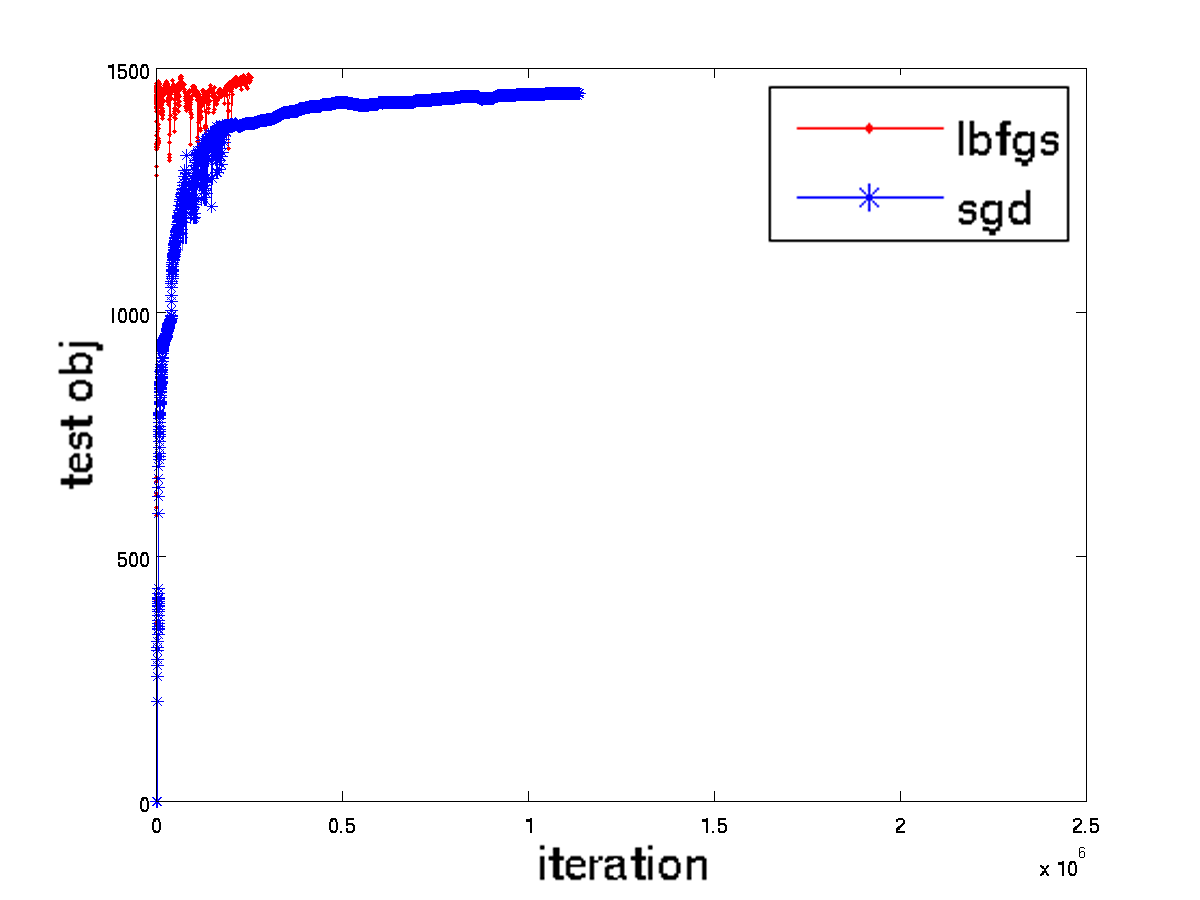} }}%
\caption{Validation error (a) and correctly classified objects as a function of the number of iterations (b) over the real-based augmented validation set. }
\label{real}
\end{figure}

\begin{figure}
\includegraphics[width=\linewidth]{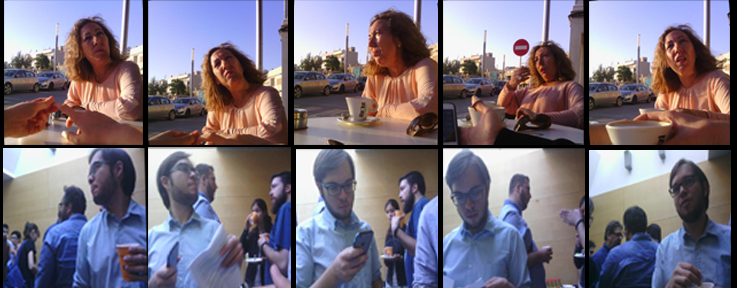}
\caption{Two examples of the final results of the method. First row classified by the model as socially interacting. Second row shows an example of non-socially interacting.}
\label{ebot}
\end{figure}

In Fig. \ref{real} (a) we report a plot of the validation error and in Fig. \ref{real} (b) we report the number of correctly classified example both as a function of the number of iterations obtained on the real-based augmented validation set. These results were obtained with a LSTM network having 35 memory cell blocks, each with 2 memory cells, and all the gates are densely connected. We have tested several configuration of networks, varying the number of hidden LSTM, and verified that a large number of memory blocks leads to overfitting, and the opposite leads to divergence. Different gates of a network (input gates, forget gates and output gates) should avoid any leakage of unwanted information flow. Hence, they should behave like gates that are either open or closed. The gates use the logistic sigmoid function $f(x) = \frac{1}{1+e^{-\alpha x}}$ as their activation function where $\alpha$ is normally set to 1. However, as mentioned by Lyu and Zhu \cite{lstm-revisit}, choosing a high value for $\alpha$ makes the gate excitation function looks more like a binary step function. We tried different values of $\alpha$ and found that bigger $\alpha$ can lead to faster and better convergence to solve our problem (3.5 in this work). Additionally, as expected an appropriately tuned batch size (500, in this case) led to the more steadier learning curve. The learning rate 0.01 and the momentum 0.8 led to the best performance of the network.

As expected, the network converges on the validation set more rapidly using LBFGS compared to SGD \cite{ngiam2011optimization}. However, SGD leads to better recall while LBFGS perform with higher precision on the test set. Overall, the SGD performs slightly better than L-BFGS comparing their F-measure.

\section{Conclusion}
This paper proposed a complete pipeline for detecting social interactions in egocentric photo-streams captured by a low frame-rate wearable camera. 
The proposed architecture consists of 2 major modules:  $1.$ frame-level social signals extraction characterizing F-formation (namely, distance and orientation of the people w.r.t. the camera wearer), and  $2.$ classification of the interaction pattern made of a two-dimensional time-series into socially interacting pattern and non-socially interacting pattern. To the best of our knowledge, this is the first time that F-formation are analyzed at sequence level instead of frame-to-frame level employing LSTM.
Experimental results have shown that the proposed method achieves high accuracy on the considered purpose (78\% using SGD). This work has important applications in the fields of preventive medicine and human-computer interaction as for example memory training by digital memories of people affected by mild cognitive impairment and robotic therapy for children affected by autism. Future work will be devoted to the characterization of the social interactions by incorporating emotion recognition from images and with the help of additional sensors.

%\section*{Acknowledgment}
%This work was partially founded by projects TIN2012-38187-C03-01 and SGR 1219. The first author is supported by an \textit{APIF} grant from the University of Barcelona.The second author is supported by a \textit{Beatriu de Pin\`os} grant (Marie-Curie COFUND action). The third author is partly supported by an \textit{ICREA Academia} grant. 

\bibliographystyle{IEEEtran}
\bibliography{biblio}

\end{document}